\documentclass[letterpaper, 10 pt, conference]{ieeeconf}  

\IEEEoverridecommandlockouts                              

\usepackage{amsmath}
\usepackage{amssymb}
\usepackage{algorithm}
\usepackage{algorithmic}
\usepackage{array}
\usepackage{adjustbox}
\usepackage{booktabs}
\usepackage{colortbl}
\usepackage{tabularx}
\usepackage{pifont}
\usepackage{graphicx}
\usepackage{multirow}
\usepackage[table]{xcolor}
\usepackage{wrapfig}
\usepackage{float}
\usepackage{hyperref}
\hypersetup{
    hidelinks
}
\usepackage{subcaption}
\usepackage{pgfplots}
\usepgfplotslibrary{polar}
\pgfplotsset{compat=1.18}
\usepackage{subcaption}
\usepackage{cuted}
\usepackage{capt-of}
\usepackage{cite}

\title{\LARGE \bf
GAE: General Action Expert for Real-Time Humanoid Teleoperation
}

\author{
    {\bfseries
  Yuefan Wang{\normalfont\textsuperscript{1 2}}
  \quad Huaicheng Zhou{\normalfont\textsuperscript{1}}
  \quad Xiao He{\normalfont\textsuperscript{1}}
  \quad Zhijie He{\normalfont\textsuperscript{1}}
  \quad Mingchuan Yang{\normalfont\textsuperscript{1}}}\\
  {\bfseries
  Huayi Zhang{\normalfont\textsuperscript{1}}
  \quad Li Chai{\normalfont\textsuperscript{2}}
  \quad Jinxin Liu{\normalfont\textsuperscript{1 *}}
  \quad Donglin Wang{\normalfont\textsuperscript{1 2 *}}
  }
  \\[4pt]
  \textsuperscript{1}Westlake Robotics \quad \textsuperscript{2}Westlake University \quad \thanks{$^{*}$Corresponding authors.}
}

\begin{document}

\maketitle
\vskip -12pt
\noindent
\makebox[\textwidth][c]{%
    \raisebox{10pt}[0pt][0pt]{%
        \textbf{Project Website: }
        \href{https://wangyf0928.github.io/gae-wlrobotics/}
        {\texttt{wangyf0928.github.io/gae-wlrobotics/}}%
    }%
}
\par\vspace*{-10pt}
\thispagestyle{empty}
\pagestyle{empty}

\begin{strip}
    \centering
    \includegraphics[width=0.9\textwidth]{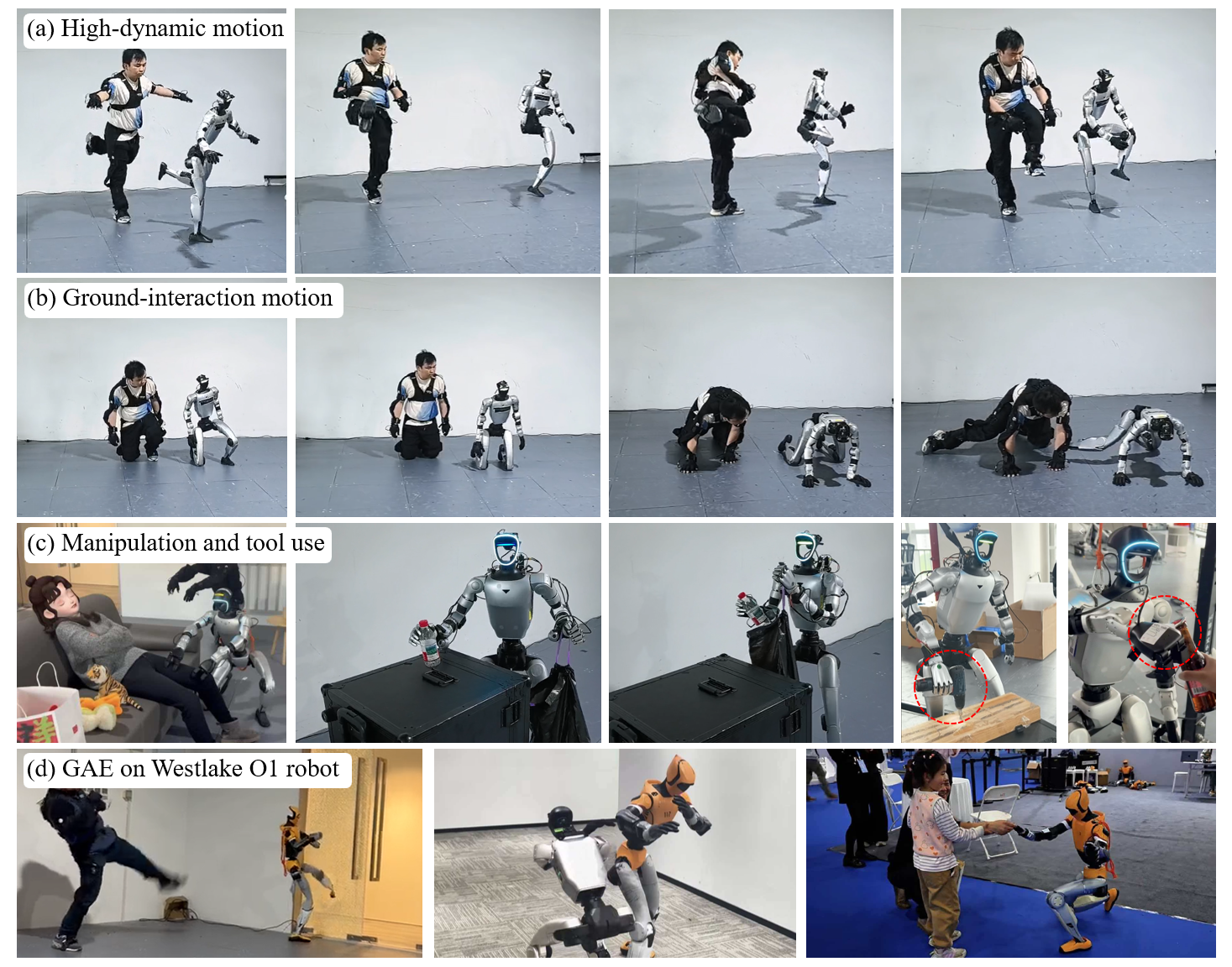}
    \captionof{figure}{
    \textbf{Real-time whole-body humanoid teleoperation with GAE.} GAE enables humanoid robots to act as responsive physical avatars of human users across diverse motions, tasks, and hardware. (a) High-dynamic motions. (b) Ground-interaction motions. (c) Manipulation and tool use with Unitree Dex3 three-finger hands and BrainCo Revo2 dexterous hands. (d) Cross-platform deployment on the Westlake O1 robot.
    }
    \label{fig:1}
\end{strip}

\begin{abstract}

Humanoid avatars extend human physical presence beyond the body, enabling people to participate in social, service, and labor activities through remotely operated robots. This requires teleoperation systems capable of realizing diverse and dynamic whole-body behaviors while maintaining responsive human-robot synchronization. We present \textbf{General Action Expert (GAE)}, a unified learning framework for general-purpose, low-latency humanoid whole-body teleoperation. To cover diverse human behaviors, GAE builds a large-scale human motion dataset from heterogeneous sources, including videos, animations, and motion capture, followed by standardization and augmentation. GAE then addresses the noise and embodiment mismatch in human motions with a two-stage training paradigm: a privileged generator policy first tracks human motion references in simulation and rolls out feasible humanoid trajectories; a deployable executor policy then learns to track these generated trajectories under curriculum domain randomization. For responsive human-robot synchronization, GAE introduces a latency-conditioned anticipation mechanism that adaptively compensates for end-to-end delay during real-time teleoperation. Simulation and real-world experiments on Unitree G1 and Westlake O1 robots demonstrate that GAE enables humanoids to smoothly mirror diverse, agile, and expressive human behaviors.

\end{abstract}

\section{INTRODUCTION}

Owing to their human-like morphology, humanoid robots are naturally compatible with human-centric environments and can reproduce a wide range of human behaviors~\cite{cheng2024expressive,chen2025gmt,wang2026omnixtreme,xie2026kungfubot}. These properties make humanoids a compelling platform for building physical avatars, where a teleoperated robot extends a user's presence and agency beyond their own body~\cite{li2026omniclone,myers2025child,zhu2026clot,darvish2023teleoperation}. Such avatars could enable people to participate in social interaction, service tasks, and physical labor in remote, hazardous, or otherwise inaccessible environments without being physically present.

Realizing this vision imposes two fundamental requirements on the teleoperation system. First, the robot must support diverse, agile, and expressive whole-body behaviors, as human operators naturally perform open-ended motions during teleoperation. Second, the robot must remain tightly synchronized with the operator, as even small delays between human motion and robot response can weaken controllability and reduce the sense of embodiment ~\cite{xiong2026extremcontrol,kamtam2024network}. 

The first requirement, broad whole-body motion coverage, has motivated recent efforts in reinforcement-learning-based humanoid control~\cite{wang2026experts}. A key prerequisite for such diversity is access to large-scale human motion data~\cite{lin2023motion,harvey2020robust}. However, high-quality motion-capture data remain limited in scale, while broader sources such as videos and character animations often contain substantial noise and ambiguity. In addition, human motions cannot be used directly as humanoid control targets due to the mismatch between human bodies and robot morphologies~\cite{araujo2025retargeting}. Kinematic retargeting can map human motions to humanoid-compatible references, but the resulting trajectories may be unstable, discontinuous, or physically inconsistent, introducing noisy learning objectives for policy. Besides, retargeting introduces an additional processing step during real-time deployment, which increases system complexity and latency~\cite{ze2025twist}. More critically, deployable policies often rely on extensive domain randomization to bridge the sim-to-real gap~\cite{zhang2025track}, which introduces complex and stochastic physical constraints into training. The compounding effect of noisy motion targets and randomized physical constraints makes policy learning unstable and difficult to scale to diverse and dynamic whole-body behaviors.

The second requirement, responsive human-robot synchronization, presents a different but equally critical challenge, yet is less commonly treated as an explicit design objective in humanoid whole-body control works~\cite{fu2024humanplus,he2024learning,he2024omnih2o}. In practical deployment, delays are inevitably introduced by sensing, communication, preprocessing, policy inference, and low-level actuation. These delays cause the robot to react to outdated human motion signals, creating a visible temporal gap between the operator and the humanoid avatar, which is particularly detrimental for motions requiring precise timing and responsiveness.

In this work, we present \textbf{General Action Expert (GAE)}, a unified learning framework for general-purpose, low-latency humanoid whole-body teleoperation. GAE targets the two challenges above: scaling humanoid control to open-ended human behaviors and maintaining responsive synchronization under real-world latency. To this end, GAE integrates three key components: a large-scale human motion dataset, a two-stage training paradigm that decouples noisy motion targets from randomized physical constraints, and a latency-conditioned anticipation mechanism.

To support broad motion diversity, GAE first builds a large-scale human motion dataset from heterogeneous sources, including videos, character animations, and motion-capture data. All motions are standardized and further expanded through three augmentation strategies: mirroring, which generates left-right variants of existing motions; stitching, which connects motion sequences with similar transitional poses; and recomposition, which recombines upper- and lower-body motions from different sequences. These procedures substantially expand the coverage of human behaviors while preserving a coherent whole-body structure, resulting in a large-scale dataset comprising more than 10,000 hours of motion data.

\begin{figure*}[t]
    \centering
    \includegraphics[width=0.95\textwidth]{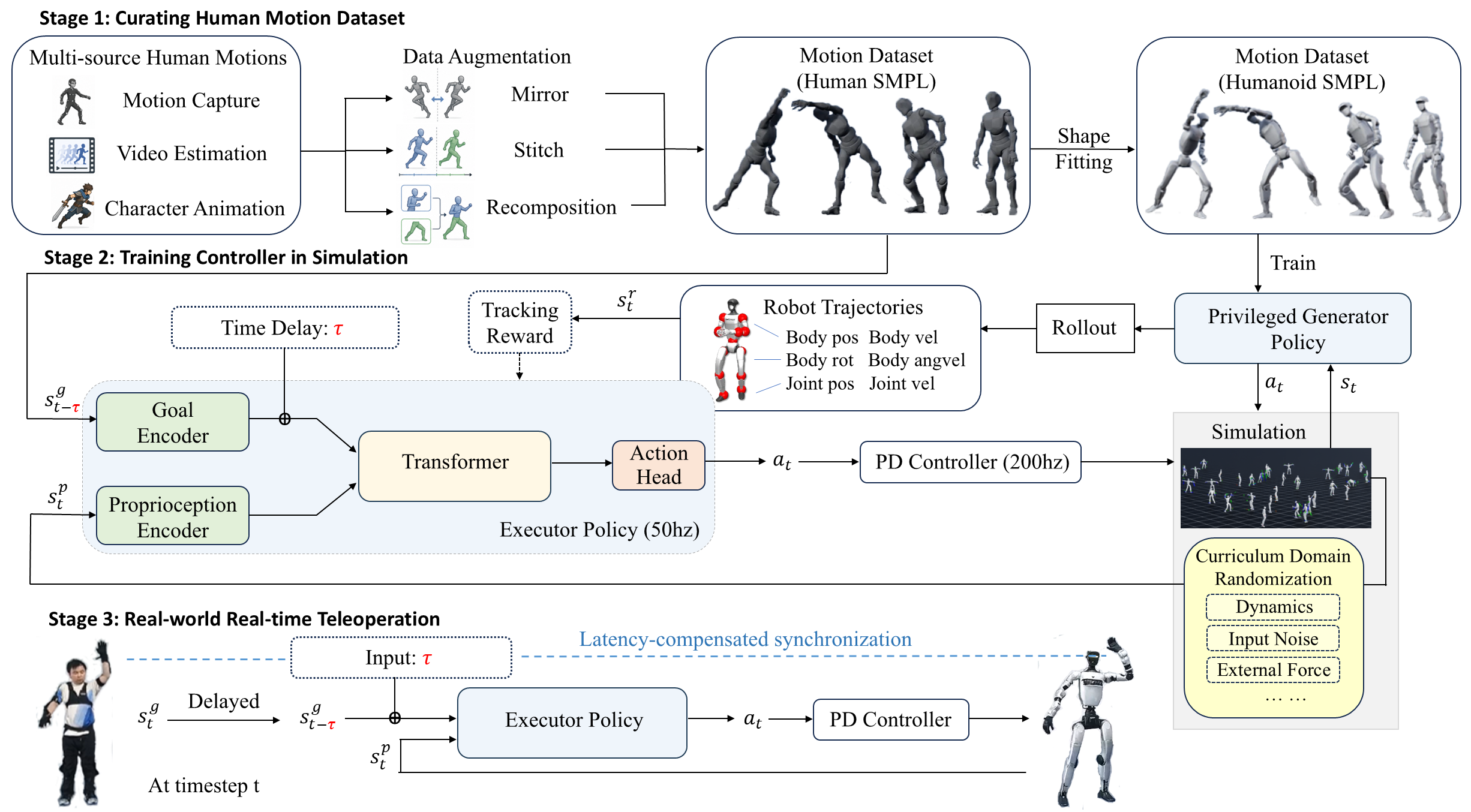}
    \caption{\textbf{Overview of GAE framework.} GAE consists of three stages: 1) curating a large-scale multi-source human motion dataset, 2) training the controller in simulation through a generator-executor paradigm, 3) teleoperating real-world robots with latency-conditioned synchronization.}
    \label{fig:framework}
    \vspace{-4pt}
\end{figure*}

On top of this data set, GAE adopts a two-stage training paradigm to enable effective large-scale policy learning for humanoid teleoperation. The key idea is to first ensure reliable tracking references and then learn sim-to-real robustness for deployment. In the first stage, human motion references are geometrically adapted to the humanoid scale through shape fitting. A privileged generator policy is then trained in a clean simulation setting to track these shape-fitted references. By rolling out this policy, GAE converts noisy and embodiment-mismatched human motions into physically feasible humanoid trajectories with accuracy and stability. In the second stage, a deployable executor policy learns to track these generated trajectories under curriculum domain randomization. 
Importantly, we explicitly condition the executor on raw human motion observations rather than the generated humanoid trajectories. As a result, the deployed policy can directly react to human motions without relying on any intermediate retargeting or trajectory-generation module. 
This design naturally separates motion denoising and embodiment adaptation from sim-to-real robustness learning. By avoiding the coupling between noisy human-motion references and randomized physical dynamics, the framework enables more stable and effective large-scale policy optimization.

To further improve teleoperation responsiveness, GAE incorporates a latency-conditioned anticipation mechanism. Instead of passively reacting to the latest available human motion, the executor learns to predictively track future human motion states based on the observed motion context. During deployment, the horizon of future anticipation can be adjusted according to the end-to-end latency of the teleoperation pipeline, allowing the policy to proactively compensate for communication and processing delays. As a result, the humanoid avatar maintains tighter temporal alignment with the operator during real-time teleoperation.

Together, these designs form a unified framework that advances humanoid teleoperation by enabling effective policy learning and latency-aware motion control. We evaluate GAE in both simulation and real-world deployment on Unitree G1 and Westlake O1 humanoid robots. As shown in Fig.~\ref{fig:1}, GAE enables humanoids to smoothly mirror diverse whole-body behaviors, including dynamic locomotion, contact-rich ground behaviors, tool-use manipulation, and loco-manipulation. These results demonstrate the potential of GAE for building responsive humanoid avatars capable of robust real-time teleoperation across diverse platforms and motion scenarios.

\section{Related Work}

\textbf{Humanoid Teleoperation.} Humanoid teleoperation offers an intuitive way to transfer human intent to robots and is essential for real-world humanoid deployment. Existing systems use various input modalities, including sparse VR keypoints, exoskeletons~\cite{zhong2025nuexo}, motion-capture suits, and camera-based pose estimation~\cite{li2026mirror}. Many of them, however, decouple upper- and lower-body control, with human commands controlling the arms or torso and a separate locomotion policy handling walking, which limits coordinated whole-body behaviors such as kicking, crouching, stepping, and legged manipulation. Recent works train policies to track human motions for whole-body teleoperation, but camera-based systems remain sensitive to occlusion and root-motion errors, while sparse VR inputs only partially observe the body. Although motion capture provides more accurate whole-body signals, it still requires a controller that can produce dynamically feasible robot actions with minimal latency. GAE addresses this by learning a unified whole-body controller directly conditioned on human motion observations, avoiding an additional online retargeting stage.

\textbf{Learning-based Humanoid Whole-body Control.} Learning-based humanoid control has evolved from task-specific locomotion policies to general whole-body motion tracking controllers. Recent works improve the generality of humanoid control by learning from large-scale human motion data~\cite{ji2024exbody2,he2025hover,yin2025unitracker,liao2025beyondmimic,luo2025sonic}. Despite some progress, robustly tracking open-ended human behaviors remains challenging. Moreover, temporal responsiveness under real-world teleoperation latency is rarely treated as an explicit controller-level objective. In contrast, our work develops a general-purpose whole-body controller that supports broader dynamic behaviors while explicitly addressing low-latency real-world teleoperation.

\section{Methods}

\subsection{Problem Formulation} 

We formulate humanoid whole-body teleoperation as a goal-conditioned reinforcement learning problem. At each control step $t$, the policy $\pi$ receives the robot proprioceptive state $s_t^p$ and a human-motion goal $s_t^g$, and outputs an action $a_t \in \mathbb{R}^{\mathrm{DOF}}$, where $\mathrm{DOF}$ denotes the number of robot joints. The action $a_t$ specifies target joint positions that are executed by a low-level PD controller. The policy is optimized with Proximal Policy Optimization (PPO)~\cite{schulman2017proximal} to maximize the expected discounted return
$\mathbb{E}\left[\sum_{t=1}^{T} \gamma^{t-1} r_t\right]$, where the reward $r_t = R(s_t^p, s_t^g)$ provides dense whole-body motion-tracking feedback. 

Unlike standard motion-tracking settings where the reference motion is assumed to be instantaneously available, real-world teleoperation inevitably introduces end-to-end latency from sensing, communication, preprocessing, policy inference, and actuation. Consequently, at time $t$, the robot receives a delayed human-motion goal $s^g_{t-\tau}$, corresponding to the motion captured $\tau$ timesteps before the current timestep. This temporal discrepancy makes it difficult to maintain synchronization between the operator and the humanoid through direct goal tracking alone. 

To address this challenge, GAE explicitly conditions the policy on the system latency $\tau$ together with the latest available human-motion goal $s^g_{t-\tau}$. Crucially, although the policy observes only the delayed goal, its tracking reward is evaluated against the current human motion $s^g_t$. In this way, the policy is trained to produce actions that best match the operator's current motion from delayed motion observations. This formulation encourages the network to internally compensate for latency, leading to tighter human-robot synchronization and more responsive teleoperation during deployment.

\subsection{Algorithm Pipeline}
\textbf{Humanoid Motion Dataset.} To support diverse and expressive whole-body teleoperation, we build a large-scale human motion dataset from multiple sources, including motion-capture datasets~\cite{AMASS}, video-estimated motions, and character animation data. We convert all motions into a unified SMPL-based human body representation and enrich the dataset through symmetry-based mirroring, temporal stitching, and upper--lower body recomposition across different motion sequences. We further construct a humanoid-proportioned SMPL~\cite{loper2023smpl} body model as a geometric proxy for the robot. The model matches the body scale and limb proportions of the humanoid robot while retaining the spherical-joint kinematic structure of SMPL, rather than directly adopting the robot's hinge-joint morphology. We then fit the collected human motions to this proxy skeleton, producing morphology-aligned motion references that provide training targets for the generator policy.

\textbf{Generator Policy Training.} The generator policy is responsible for producing physically feasible robot motion trajectories with accuracy and stability that serve as tracking target for an executor policy. During training, we learn the generator policy $\pi^{\text{gen}}(\cdot \mid s_t^{\text{p-gen}}, s_t^{\text{g-gen}})$ in a clean simulation setting, \textit{i.e.,} it has access to privileged information and is not exposed to any environment disturbances. Formally, the proprioception state is defined as  $s_t^{\text{p-gen}} \triangleq [\mathbf{p}_{t}^{\text{loc}}, \dot{\mathbf{p}}_{t}, \boldsymbol{\omega}_{t}, \mathbf{f}_{t}, \mathbf{r}_{t}, \boldsymbol{\omega}_{t}^{\text{root}}, \mathbf{q}_{t}, \dot{\mathbf{q}}_{t}, \mathbf{a}_{t-1}]$, which contains the local body position $\mathbf{p}_{t}^{\text{loc}}$, linear velocity $\dot{\mathbf{p}}_{t}$, angular velocity $\boldsymbol{\omega}_{t}$, contact force $\mathbf{f}_{t}$, root orientation $\mathbf{r}_{t}$, root angular velocity $\boldsymbol{\omega}_{t}^{\text{root}}$, joint position $\mathbf{q}_{t}$, joint velocity $\dot{\mathbf{q}}_{t}$, and the previous actions $\mathbf{a}_{t-1}$. The goal state is defined as $s_t^{\text{g-gen}} \triangleq [\hat{\mathbf{p}}_{t}^{\text{loc}}, \hat{\mathbf{v}}_{t}, \hat{\mathbf{r}}_{t}]$, which contains local body positions, linear velocities, and body rotations of reference motion. The complete observation provided to the generator policy consists of a 100-frame history of proprioceptive and goal states, together with a 32-frame future of goal states. After training the generator policy, we collect robot motion trajectories by rolling it out in simulation. Each trajectory records the robot's body states $\{s^r_t\}$, including body positions, linear/angular velocities, rotations, joint positions, and joint velocities.

\textbf{Executor Policy Training.} Given the clean and feasible motion targets $\{s^r_t\}$ collected from the generator policy, the executor policy aims to learn a general and responsive controller that can be directly deployed for real-world humanoid teleoperation. To bridge the sim-to-real gap, we train the executor policy with curriculum domain randomization, where the strength and diversity of domain
randomization are progressively increased during training. This curriculum gradually exposes the policy to variations in robot dynamics, observation noise, and external perturbations, allowing it to first acquire stable motion-tracking behaviors and then improve robustness under increasingly challenging conditions. Importantly, the executor policy observes motion targets constructed from raw human motions, rather than from the generator trajectories. The generator trajectories are used only for reward computation to shape the executor's learning objective. Consequently, at deployment time, human motions can be directly used as inputs to the executor policy, without requiring an intermediate retargeting step to robot trajectories. Formally, the proprioception state of the executor policy is defined as $s_t^{\text{p}} \triangleq [\mathbf{g}_{t}, \boldsymbol{\omega}_{t}^{\text{root}}, \mathbf{q}_{t}, \dot{\mathbf{q}}_{t}, \mathbf{a}_{t-1}]$, which includes 100-frame history of robot state. The goal state is $s_t^{\text{g}} \triangleq 
[\hat{\mathbf{p}}_{t}^{\text{loc}}, \hat{\mathbf{v}}_{t}, \hat{\mathbf{r}}_{t}]$, which is represented in a format similar to that used by the generator policy. The reward is computed based on the rollout state $s_t^r$ recorded in the generated trajectory.

\textbf{Latency-conditioned Anticipation.} To further improve real-time synchronization, we introduce a latency-conditioned anticipation mechanism for the executor policy. In practical teleoperation, the human motion received by the robot is inevitably delayed. To compensate for this delay, we explicitly condition the executor policy $\pi^{\text{exe}}(\cdot \mid s_t^{\text{p}}, s_{t-\tau}^{\text{g}}, \tau)$ on a time delay $\tau$, allowing it to track human motion states predictively rather than mimicking the delayed motion observation itself. Specifically, we tokenize the raw human-motion observations together with the robot proprioceptive observations and encode their temporal positions using rotary position embeddings (RoPE)~\cite{su2024roformer}. The time delay, which we also refer to as the policy anticipation horizon, is represented by shifting the RoPE indices of the human-motion tokens. This shift changes the temporal motion frame interpreted by the policy: a larger anticipation horizon indicates that the policy should track a farther future motion state based on the current human-motion observations. During training, the anticipation horizon is randomly sampled across parallel environments, so that a single executor policy can adapt to different deployment delays. During deployment, this horizon can be adjusted according to the measured latency of the teleoperation pipeline.

\subsection{Training Details}

\textbf{Network Architecture.} Both the generator policy and the executor policy use two separate MLP encoders to encode the proprioceptive state and the goal state, respectively. The encoded features are then processed by a mixture of expert Transformer~\cite{NIPS2017_transformers} backbone, and the temporal position embeddings are applied in every block of the backbone. Finally, an MLP action head maps the last output token to the action $a_t$. Each policy contains approximately 852M parameters. We train our policies in Isaac Lab~\cite{mittal2025isaaclab} with parallel environments.

\textbf{Reward Design and Domain Randomization.} The generator policy and executor policy share a similar reward design, differing primarily in their reference motions. For the generator policy, the tracking reward is computed with respect to the shape-fitted human motion, while for the executor policy, the tracking reward is computed with respect to the trajectories rolled out by the generator policy. We formulate the reward $r_t$ as the sum of two components: 1) penalty, and 2) task rewards for motion tracking. We apply extensive curriculum domain randomization on the executor policy training for successful sim-to-real transfer. The details of the parameters for reward designs and domain randomization are provided in the Appendix~\ref{sec:appendix-training}.

\section{Experiments}

In this section, we conduct extensive experiments to demonstrate that:
\textbf{(1)} GAE accurately tracks diverse and expressive human motions;
\textbf{(2)} GAE maintains strong tracking performance across different anticipation horizons for latency compensation;
\textbf{(3)} GAE enables robust real-time teleoperation in open-world scenarios; and
\textbf{(4)} GAE training framework can be transferred across different humanoid embodiments.

\begin{table}[t]
\centering
\caption{Quantitative results.}
\label{tab:sim-results}
\scriptsize
\setlength{\tabcolsep}{2.5pt}

\makebox[\columnwidth][l]{%
\resizebox{1.01\columnwidth}{!}{%
\begin{tabular}{lccccc}
\toprule
Method & SR (\%) $\uparrow$ & MKPE $\downarrow$ & MKLVE $\downarrow$ & MKAVE $\downarrow$ & MKRE $\downarrow$ \\
\midrule
\rowcolor{gray!10}
\multicolumn{6}{l}{(a) Comparison with SONIC on Different Test Sets} \\
\multicolumn{6}{l}{\textit{Easy Set}} \\
SONIC & 95.4 & 0.036 & 0.271 & \textbf{0.923} & \textbf{0.224} \\
GAE & \textbf{97.3} & \textbf{0.033} & \textbf{0.244} & 1.382 & 0.275 \\
\addlinespace[2pt]
\multicolumn{6}{l}{\textit{Medium Set}} \\
SONIC & 85.2 & \textbf{0.045} & 0.437 & 1.385 & 0.278 \\
GAE & \textbf{96.9} & 0.046 & \textbf{0.263} & \textbf{1.076} & \textbf{0.269} \\
\addlinespace[2pt]
\multicolumn{6}{l}{\textit{Hard Set}} \\
SONIC & 77.7 & \textbf{0.043} & 0.595 & 1.489 & 0.367 \\
GAE & \textbf{93.9} & 0.045 & \textbf{0.241} & \textbf{1.443} & \textbf{0.347} \\
\midrule
\rowcolor{gray!10}
\multicolumn{6}{l}{(b) Performance under Different Anticipation Horizons} \\
$\tau=0$ & 93.9 & 0.045 & 0.241 & 1.443 & 0.347 \\
$\tau=1$ & 93.7 & 0.046 & 0.252 & 1.512 & 0.355 \\
$\tau=2$ & 94.0 & 0.048 & 0.270 & 1.598 & 0.365 \\
$\tau=3$ & 93.8 & 0.050 & 0.290 & 1.692 & 0.376 \\
$\tau=4$ & 92.8 & 0.053 & 0.309 & 1.764 & 0.389 \\
$\tau=5$ & 92.1 & 0.056 & 0.326 & 1.837 & 0.403 \\
\bottomrule
\end{tabular}
}%
}

\vspace{-2mm}
\end{table}

\textbf{Experimental Setup and Metrics.} To evaluate the model under different levels of motion complexity, we construct three evaluation motion sets, denoted as Easy, Medium, and Hard, each containing approximately 50 motion sequences that are excluded from the training set. The Easy Set consists of low-dynamic motions such as walking and waving, the Medium Set includes more dynamic motions such as running, kicking, and squatting, while the Hard Set contains highly dynamic and challenging motions such as jumping, crawling and standing up. These three sets exhibit progressively increasing motion dynamics and are used to assess the model's robustness and generalization across different levels of motion difficulty. We report four tracking errors over selected body keypoints: root-relative Mean Keypoint Position Error \textbf{(MKPE)}, Mean Keypoint Rotation Error \textbf{(MKRE)}, Mean Keypoint Linear Velocity Error \textbf{(MKLVE)}, and Mean Keypoint Angular Velocity Error \textbf{(MKAVE)}. These metrics measure whether the robot can accurately reproduce the reference motion in both pose and velocity space. In addition, we report a motion tracking Success Rate \textbf{(SR)}. A motion sequence is considered successfully tracked only if every tracked keypoint remains within $50$ cm of the reference trajectory. Compared with average tracking errors, SR better captures complete-sequence robustness by penalizing transient large deviations that may cause loss of synchronization.

\definecolor{ourscolor}{RGB}{103,157,183}     
\definecolor{basecolor}{RGB}{167,217,233}     
\definecolor{weakcolor}{RGB}{231,199,190}     
\definecolor{ourstext}{RGB}{69,118,142}      
\definecolor{basetext}{RGB}{82,145,170}      
\definecolor{weaktext}{RGB}{166,119,107}     
\definecolor{gridcolor}{RGB}{220,220,220}

\begin{figure}[t]
    \raggedleft

    \begin{subfigure}[t]{\columnwidth}
        \vspace{0pt}
        \centering

        \begin{tikzpicture}
        \begin{polaraxis}[
            width=0.576\columnwidth,
            height=0.576\columnwidth,,
            scale only axis,
            ymin=0,
            ymax=1.0,
            xtick={90,18,306,234,162},
            xticklabels={{SR},{MKPE},{MKLVE},{MKAVE},{MKRE}},
            ytick={0.4,0.6,0.8,1.0},
            yticklabels={},
            grid=none,
            axis line style={draw=none},
            tick style={draw=none},
            xticklabel style={
                font=\scriptsize,
                inner sep=1pt
            },
            legend style={
            at={(0.999,0.999)},
            anchor=north east,
            legend columns=1,
            draw=none,
            fill=white,
            fill opacity=0.85,
            text opacity=1,
            font=\tiny,
            row sep=1pt
        }
        ]
        \addplot[
            draw=gray!35,
            thin,
            forget plot
        ]
        coordinates {
            (90,0.2)
            (18,0.2)
            (-54,0.2)
            (-126,0.2)
            (162,0.2)
            (90,0.2)
        };
        
        \addplot[
            draw=gray!35,
            thin,
            forget plot
        ]
        coordinates {
            (90,0.4)
            (18,0.4)
            (-54,0.4)
            (-126,0.4)
            (162,0.4)
            (90,0.4)
        };
        
        \addplot[
            draw=gray!35,
            thin,
            forget plot
        ]
        coordinates {
            (90,0.6)
            (18,0.6)
            (-54,0.6)
            (-126,0.6)
            (162,0.6)
            (90,0.6)
        };
        
        \addplot[
            draw=gray!35,
            thin,
            forget plot
        ]
        coordinates {
            (90,0.8)
            (18,0.8)
            (-54,0.8)
            (-126,0.8)
            (162,0.8)
            (90,0.8)
        };
        
        \addplot[
            draw=gray!35,
            thin,
            forget plot
        ]
        coordinates {
            (90,1.0)
            (18,1.0)
            (-54,1.0)
            (-126,1.0)
            (162,1.0)
            (90,1.0)
        };
        
        \addplot[draw=gray!35, thin, forget plot]
        coordinates {(90,0) (90,1.0)};
        
        \addplot[draw=gray!35, thin, forget plot]
        coordinates {(90,0) (90,1.0)};
        
        \addplot[draw=gray!35, thin, forget plot]
        coordinates {(18,0) (18,1.0)};
        
        \addplot[draw=gray!35, thin, forget plot]
        coordinates {(-54,0) (-54,1.0)};
        
        \addplot[draw=gray!35, thin, forget plot]
        coordinates {(-126,0) (-126,1.0)};
        
        \addplot[draw=gray!35, thin, forget plot]
        coordinates {(162,0) (162,1.0)};
        \addplot[
            ourscolor,
            very thick,
            mark=*,
            mark size=1.4pt
        ]
        coordinates {
            (90,0.85)
            (18,0.93)
            (-54,0.830)
            (-126,0.832)
            (162,0.81)
            (90,0.85)
        };
        \addlegendentry{Full + Aug.}

        \node[ourstext,font=\tiny,yshift=5pt]
            at (axis cs:90,0.85) {93.9};
        \node[ourstext,font=\tiny,xshift=7pt,yshift=3pt]
            at (axis cs:22,0.82) {0.045};
        \node[ourstext,font=\tiny,xshift=7pt,yshift=-3pt]
            at (axis cs:-54,0.840) {0.241};
        \node[ourstext,font=\tiny,xshift=-2pt,yshift=-7pt]
            at (axis cs:-126,0.832) {1.443};
        \node[ourstext,font=\tiny,xshift=-8pt,yshift=2pt]
            at (axis cs:162,0.81) {0.347};

        \addplot[
            basecolor,
            very thick,
            mark=square*,
            mark size=1.3pt
        ]
        coordinates {
            (90,0.79)
            (18,0.78)
            (-54,0.738)
            (-126,0.768)
            (162,0.74)
            (90,0.79)
        };
        \addlegendentry{Full w/o Aug.}

        \node[basetext,font=\tiny,xshift=-10pt,yshift=1pt]
            at (axis cs:90,0.79) {91.6};
        \node[basetext,font=\tiny,xshift=7pt,yshift=1pt]
            at (axis cs:14,0.8) {0.048};
        \node[basetext,font=\tiny,xshift=7pt,yshift=-2pt]
            at (axis cs:-54,0.568) {0.271};
        \node[basetext,font=\tiny,xshift=-2pt,yshift=-7pt]
            at (axis cs:-126,0.578) {1.562};
        \node[basetext,font=\tiny,xshift=-8pt,yshift=1pt]
            at (axis cs:168,0.74) {0.367};

        \addplot[
            weakcolor,
            very thick,
            mark=triangle*,
            mark size=1.5pt
        ]
        coordinates {
            (90,0.63)
            (18,0.59)
            (-54,0.446)
            (-126,0.620)
            (162,0.54)
            (90,0.63)
        };
        \addlegendentry{50\% w/o Aug.}

        \node[weaktext,font=\tiny,xshift=8pt,yshift=1pt]
            at (axis cs:90,0.63) {85.3};
        \node[weaktext,font=\tiny,xshift=7pt,yshift=1pt]
            at (axis cs:15,0.59) {0.053};
        \node[weaktext,font=\tiny,xshift=7pt,yshift=-2pt]
            at (axis cs:-54,0.446) {0.448};
        \node[weaktext,font=\tiny,xshift=-1pt,yshift=-7pt]
            at (axis cs:-126,0.40) {1.936};
        \node[weaktext,font=\tiny,xshift=-8pt,yshift=1pt]
            at (axis cs:162,0.54) {0.446};
        \end{polaraxis}
        \node[ anchor=north west, font=\small\bfseries ] at ([xshift=2pt,yshift=-2pt]current axis.north west) {(a)};
        \end{tikzpicture}
        \vspace{3mm}
        \label{fig:data_radar}
    \end{subfigure}
    \vspace{1mm}
    \begin{subfigure}[t]{\columnwidth}
        \vspace{0pt}
        \centering

        \begin{tikzpicture}
        \begin{axis}[
            width=0.74\columnwidth,
            height=2.2cm,
            scale only axis,
            ybar,
            bar width=13pt,
            ymin=0.86,
            ymax=1.0,
            enlarge x limits={abs=1pt},
            symbolic x coords={LEFT,Easy Set,Medium Set,Hard Set,RIGHT},
            xmin=LEFT,
            xmax=RIGHT,
            xtick=data,
            xticklabel style={
                font=\scriptsize,
                rotate=25,
                anchor=east
            },
            ylabel={SR (\%)},
            ylabel style={
                font=\scriptsize,
                yshift=-2pt
            },
            ytick={},
            yticklabels={},
            grid=none,
            axis x line=bottom,
            axis y line=left,
            axis line style={
                draw=black!60,
                thin
            },
            tick style={
                draw=none
            },
            legend style={
                at={(0.999,0.999)},
                anchor=north east,
                legend columns=1,
                draw=none,
                fill=white,
                fill opacity=0.85,
                text opacity=1,
                font=\tiny,
                row sep=1pt
            },
            nodes near coords,
            every node near coord/.append style={
                font=\tiny,
                rotate=90,
                anchor=west,
                yshift=1pt
            },
            point meta=explicit symbolic
        ]

        \addplot[
            fill=ourscolor,
            draw=ourscolor!80!black,
            line width=0.45pt,
            bar shift=-15pt
        ]
        coordinates {
            (Easy Set,0.973)    [97.3]
            (Medium Set,0.969)  [96.9]
            (Hard Set,0.939) [93.9]
        };
        \addlegendentry{Generator-executor}

        \addplot[
            fill=basecolor,
            draw=basecolor!80!black,
            line width=0.45pt,
            bar shift=1pt
        ]
        coordinates {
            (Easy Set,0.969)    [96.9]
            (Medium Set,0.954)  [95.4]
            (Hard Set,0.906) [90.6]
        };
        \addlegendentry{Train-from-scratch}
        \end{axis}
        \node[ anchor=north west, font=\small\bfseries ] at ([xshift=2pt,yshift=-2pt]current axis.north west) {(b)};
        \end{tikzpicture}

        \label{fig:two_stage_ablation}
    \end{subfigure}

    \caption{
        Ablation studies on data composition and training strategy.
        (a) Effect of data scaling and augmentation.
        (b) Comparison between our generator-executor paradigm and train-from-scratch.
    }
    \label{fig:training_ablation}
    \vspace{-6pt}
\end{figure}
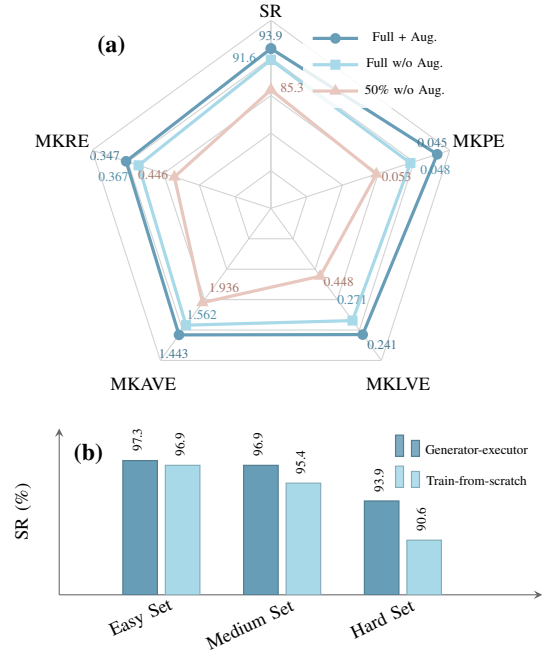

\begin{figure*}[t]
    \centering
    \includegraphics[width=0.98\textwidth]{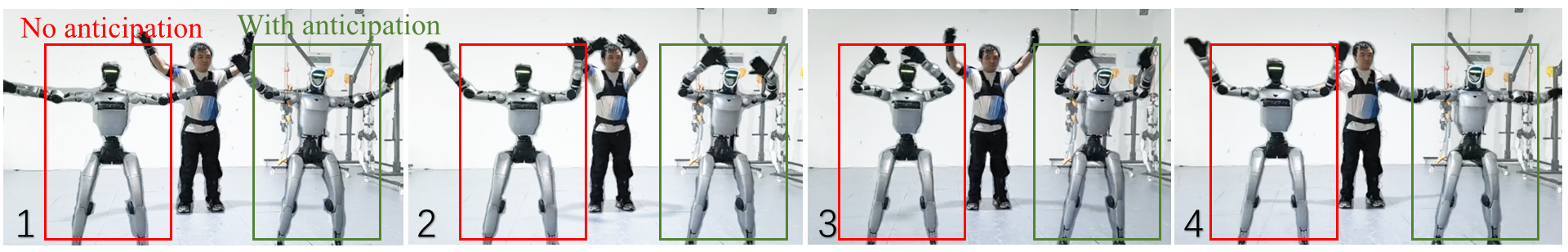}
    \caption{Synchronization evaluation in real world.}
    \label{fig:synchronization-real}
    \vspace{-2pt}
\end{figure*}

\subsection{Simulation Motion Tracking Results}

We first compare GAE with SONIC, a representative and currently state-of-the-art open-source whole-body controller for humanoid robots. As shown in Table ~\ref{tab:sim-results}(a), GAE achieves performance comparable to SONIC on the Easy set, where both methods can reliably track relatively simple whole-body motions. The advantage of GAE becomes more pronounced as the motion difficulty increases. GAE consistently achieves better overall performance on the Medium and Hard sets. In particular, GAE maintains higher sequence-level success rates as motion dynamics increase, showing stronger robustness for challenging whole-body motions. These results indicate that GAE can reliably support a broader range of human motions, especially highly dynamic ones.

We further evaluate whether GAE can support deployment-time latency compensation through adaptive anticipation horizons. Since the controller runs at $50$ Hz, advancing the tracking target by one frame corresponds to $20$ ms of anticipation. Thus, anticipation horizons from $\tau=0$ to $\tau=5$ provide $0$--$100$ ms of lead time, enabling the same policy to compensate for system latency up to $100$ ms. However, anticipating further into the future makes motion tracking more challenging and may reduce tracking accuracy. To study this trade-off, we evaluate the same trained GAE model under different anticipation horizons on the hard set, where highly dynamic motions impose the greatest challenge for anticipation-based control. Importantly, no model is retrained for different horizons; we only adjust the horizon at evaluation time through the temporal position embedding.

As shown in Table~\ref{tab:sim-results}(b), increasing the anticipation horizon leads to a gradual but moderate degradation in tracking precision. As $\tau$ increases from $0$ to $5$, MKPE increases from $0.045$ to $0.056$, and the velocity- and rotation-related errors also grow consistently. This trend is expected, since a larger horizon requires the policy to track farther states and makes anticipation-based control more challenging, especially on the hard set with highly dynamic motions. Nevertheless, the overall performance remains stable across all horizons. Even with $\tau=5$, which corresponds to $100$ ms of anticipation, GAE still achieves a success rate of $92.1\%$ and maintains relatively low tracking errors. These results indicate that GAE can trade a small amount of tracking precision for deployment-time latency compensation, while still robustly executing challenging whole-body motions.

\textbf{Ablation study.} We conduct ablation studies to investigate the effects of data composition and training strategy. As shown in Fig.~\ref{fig:training_ablation}(a), increasing the amount of training data improves the overall tracking performance, while our data augmentation strategy provides further gains, demonstrating that large data scale and appropriate augmentation strategy are effective for learning a general controller. Fig.~\ref{fig:training_ablation}(b) compares our generator–executor training paradigm with directly training the policy from scratch. Our approach consistently achieves higher success rates across all three evaluation sets, with the advantage becoming more pronounced as motion difficulty increases. We attribute this improvement to our two-stage formulation, which decouples noisy human-motion references from randomized physical dynamics, thereby reducing the optimization difficulty and leading to better robustness on challenging motions.

\subsection{Real-world Motion Tracking Results}

\textbf{Latency compensation through anticipation.} In real-world deployment, we first evaluate whether the anticipation mechanism of GAE can reduce the visible delay between the human operator and the robot. As shown in Fig.~\ref{fig:synchronization-real}, two Unitree G1 robots are simultaneously controlled by GAE using $\tau=0$ and $\tau=4$, respectively. The latter corresponds to an $80$ ms anticipation horizon at our $50$ Hz control frequency. The comparison shows a clear difference in motion synchronization. Without anticipation, the robot exhibits an observable lag behind the operator: when the operator has already moved to a new pose, the robot is still reproducing an earlier pose. In contrast, the robot using $\tau=4$ remains more synchronized with the human motion throughout the sequence. This result demonstrates the effectiveness of the proposed anticipation mechanism, showing that explicitly conditioning the policy on the anticipation horizon improves temporal synchronization between the operator and the humanoid robot.

\textbf{Real-world teleoperation.}
We further evaluate GAE through extensive real-world teleoperation experiments covering diverse whole-body behaviors, as summarized in Fig.~\ref{fig:1}. Beyond regular locomotion, GAE supports highly dynamic motions such as rapid stepping, kicking, single-leg support and large-amplitude whole-body movements, as well as challenging ground-interaction motions involving kneeling, crouching, and crawling. GAE also enables stable upper-body manipulation and tool-use behaviors when integrated with dexterous hands. These experiments involve diverse and continuously varying behaviors, highlighting the robustness and versatility of GAE for real-time physical-avatar control in practical settings.

\textbf{Transfer to other humanoid embodiments.}
We also investigate whether the GAE framework can generalize beyond the Unitree G1 embodiment. As shown in Fig.~\ref{fig:1}(d), we transfer the framework to the Westlake O1 robot, whose morphology, joint configuration, and hardware characteristics differ from those of G1. After embodiment-specific fine-tuning, the resulting controller enables O1 to reproduce diverse human whole-body motions, including dynamic movements and interactive behaviors. Together with the extensive demonstrations on G1, these results indicate that GAE is not tied to a specific humanoid platform, but provides a general framework that can be adapted to different robot embodiments for real-time control.

\section{Discussion}

\subsection{Limitations and Future Work} 
Although GAE enables general and low-latency whole-body teleoperation, the current system still lacks explicit perception of the surrounding environment. The controller primarily relies on human motion targets and robot proprioceptive states, without incorporating external observations such as RGB images or LiDAR. Consequently, it cannot directly reason about environmental geometry or proactively adapt its behavior to terrain and obstacles, which limits its applicability to more challenging scenarios such as stair climbing, obstacle traversal, and other contact-rich interactions with the environment. In future work, we plan to incorporate richer exteroceptive perception into GAE, enabling the controller to jointly reason about motion targets, robot states, and environmental geometry. We expect this extension to further improve the generality and robustness of whole-body teleoperation in complex and unstructured environments.

\subsection{Interesting Findings}
GAE is originally designed as a motion-tracking policy that follows externally provided human motion targets. Interestingly, we find that meaningful motion can sometimes be sustained even when the external target is removed. Specifically, instead of feeding the policy a new human motion target at every step, we use forward kinematics (FK) to convert the robot’s own state into the same motion target format and feed it back as the target for the subsequent step, forming a simple self-referential closed loop:
\[
\begin{aligned}
\mathrm{(I)}\qquad
&\bigl(\text{robot state},\; \text{human motion target}\bigr)
\xrightarrow{\mathrm{GAE}} \text{action}, \\[2pt]
\mathrm{(II)}\qquad
&\bigl(\text{robot state},\; \mathrm{FK}(\text{robot state})\bigr)
\xrightarrow{\mathrm{GAE}} \text{action}. \\[2pt]
&\mathrm{(I)}:\ \text{Teleoperation};\qquad
\mathrm{(II)}:\ \text{Self-referential}.
\end{aligned}
\]

\begin{figure}[H]
    \centering
    \includegraphics[width=\linewidth]{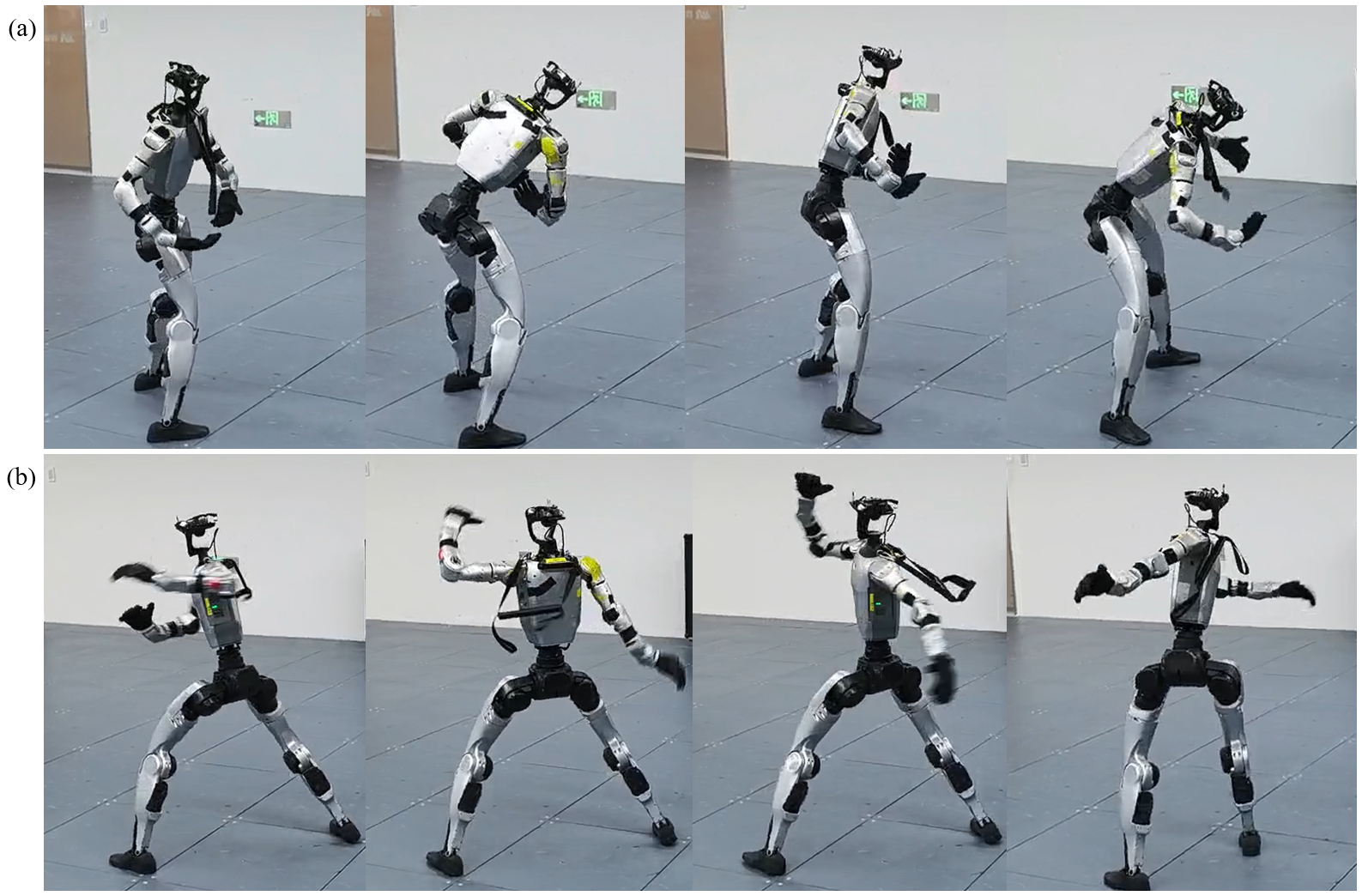}
    \caption{"free-style" produced by self-referential control. (a) Cradling sway; (b) Wide arm swings.}
    \vspace{-4pt}
    \label{fig:freestyle}
\end{figure}

At first glance, one might expect such a system to gradually converge to a static configuration. If the target is always derived from the robot’s current state, the seemingly trivial solution would be to simply remain where it is. Surprisingly, this is not always what we observe. When the robot is already engaged in a structured dynamic behavior, such as dancing or walking, the closed-loop system can continue to sustain and propagate the ongoing motion over time. Walking may settle into a persistent locomotion cycle, while a dancing sequence may evolve into the robot's own "free-style" behavior, forming a self-sustained motion loop that repeatedly generates coherent dance-like movements without an external motion reference, as shown in Fig. \ref{fig:freestyle}.

This observation suggests that GAE may have learned more than a simple reactive rule of “given a target pose, track that pose.” Instead, the policy appears to implicitly capture certain motion priors or dynamics priors about how physically meaningful behaviors evolve over time. For example, it may encode regularities such as what typically follows the current phase of a walking cycle, how the current velocity, body configuration, and joint state constrain the subsequent motion, and which state transitions correspond to dynamically feasible and sustainable behaviors.

Consequently, even when the input does not explicitly specify what the next motion should be, the combination of the robot’s current state and the motion structure embedded in the learned policy can be sufficient to maintain an ongoing dynamic pattern. In this sense, the robot is not merely tracking an externally specified trajectory; the closed-loop system is able to preserve and propagate part of the temporal structure already present in its current behavior.

A useful way to interpret this phenomenon is through the closed-loop dynamics induced jointly by the policy, the robot dynamics, and the self-referential signal. Certain behaviors, such as walking, may correspond to stable dynamic modes or limit-cycle-like attractors of this closed-loop system. Once the robot enters such a mode, the feedback loop can keep the state evolving along a locomotion manifold rather than causing it to immediately collapse to a static equilibrium.

Importantly, this behavior should not be interpreted as evidence that GAE performs explicit motion planning or fully autonomous motion generation. Rather, it suggests that a sufficiently general motion-tracking policy may internally compress part of the temporal and dynamical structure of natural motion. From this perspective, GAE is not only passively following a reference; it appears to have learned, to some extent, how motion naturally evolves over time.

\section*{ACKNOWLEDGMENT}

We would like to thank all members of Westlake Robotics and the MiLAB from Westlake University for their support. This work was supported by the Brain Science and Brain-like Intelligence Technology — National Science and Technology Major Project (Grant No. 2022ZD0208800)


\bibliographystyle{IEEEtran}
\bibliography{references}
\clearpage
\section*{APPENDIX}
\setcounter{subsection}{0}
\renewcommand{\thesubsection}{\Alph{subsection}}

\subsection{Training Details}
\label{sec:appendix-training}

\textbf{Reward Design.} In Table \ref{tab:reward-design}, we summarize the reward terms used for RL training, including penalty terms for regularization and safety, as well as task rewards for whole-body motion tracking. The corresponding weights are selected to balance tracking accuracy, smoothness, and stability.




\begin{table}[htbp]
\centering
\caption{Reward designs for RL training.}
\label{tab:reward-design}
\small
\setlength{\tabcolsep}{1pt}
\renewcommand{\arraystretch}{1.2}
\begin{tabular*}{\columnwidth}{@{\extracolsep{\fill}}lcc@{}}
\toprule
\textbf{Term} & \textbf{Equation} & \textbf{Weight} \\
\midrule
\multicolumn{3}{l}{\textbf{Penalty}}\\[-2pt]
\addlinespace[2pt]

Torque limits
& $\sum_j [|\tau_j^{\rm c}/\bar\tau_j|-0.75]_+$
& $-1.0$ \\
DoF position limits
& $\sum_j ([q_j^{min} -q_j]_+ +[q_j-q_j^{max}]_+)$
& $-1.0\!\times\!10^{1}$ \\
DoF velocity limits
& $\sum_j [|\dot q_j/\bar v_j|-\rho]_+$
& $-1.0$ \\
Torque
& $\sum_j |\beta_j^\tau\tau_j^{\rm c}|$
& $-1.0\!\times\!10^{-4}$ \\
Action rate
& $\|\mathbf a_t-\mathbf a_{t-1}\|_2^2$
& $-1.0\!\times\!10^{-2}$ \\
DoF acceleration
& $\sum_j |\ddot q_j|$
& $-5.0\!\times\!10^{-5}$ \\
DoF velocity
& $\sum_j \dot q_j^{\,2}$
& $-1.0\!\times\!10^{-3}$ \\
Power
& $\sum_j |\beta_j^P\tau_j^{\rm a}\dot q_j|$
& $-1.0\!\times\!10^{-4}$ \\
\midrule
\addlinespace[2pt] \multicolumn{3}{l}{\textbf{Task Reward}}\\[-2pt] 
\addlinespace[2pt]
Body pos tracking
& $\exp(-k_p\sum_b\bar w_b^p\|\hat{\mathbf p}_b^{\rm rel}-\mathbf p_b^{\rm rel}\|_2)$
& $1.0$ \\
Body vel tracking
& $\exp(-k_v\sum_b\bar w_b^v\|\mathbf D(\hat{\mathbf v}_b-\mathbf v_b)\|_2)$
& $1.0$ \\
Body ang-vel tracking
& $\exp(-k_\omega\sum_b\bar w_b^\omega\|\hat{\boldsymbol\omega}_b-\boldsymbol\omega_b\|_2)$
& $1.0$ \\
Body rot tracking
& $\exp(-k_R\sum_b\bar w_b^R\theta_b)$
& $1.0$ \\
\bottomrule
\end{tabular*}
\vspace{3pt}

\begin{minipage}{\columnwidth}
\small
\textit{Notation.}
$[x]_+=\max(x,0)$; $\tau^{\rm c}$ and $\tau^{\rm a}$ denote computed and applied torques;
$\beta^\tau,\beta^P$ are joint multipliers ($w$ on selected joints,
$1$ otherwise). $\rho$ is the soft ratio of joint velocity limit.
For tracking, hats denote references, and
\[
\begin{aligned}
\mathbf p_b^{\rm rel}&=\mathbf p_b-\mathbf p_{\rm root},
&\bar w_b^x&=\frac{w_b^x}{\sum_{i\in\mathcal B_x}w_i^x},\\
\mathbf D&=\operatorname{diag}(1,1,2),
&\theta_b&=2\arccos\!\left(|\hat Q_b^\top Q_b|\right).
\end{aligned}
\]
$k_x,s_x$ are sensitivity and scale.
$Q_b,\hat Q_b$ are orientation quaternions;
$\mathbf v_b,\boldsymbol\omega_b$ are world-frame velocities. Tracking sums run over the selected body links $\mathcal B_x$, including pelvis, torso, hip, knee, ankle, shoulder, elbow, and wrist. 
\end{minipage}
\end{table}

\textbf{Auxiliary Objectives.} We introduce two auxiliary objectives to further improve the representation capability of the policy. Specifically, two MLP heads are attached to the shared Transformer backbone to predict the future motion target and future robot state, respectively. Future-target prediction encourages the backbone to capture the temporal evolution of the motion target, while future-state prediction provides additional supervision for modeling the robot's state transition and dynamics. The total loss comprises:

\[
\mathcal{L}
=
\mathcal{L}_{\mathrm{PPO}}
+
\lambda_{\mathrm{t}} \mathcal{L}_{\mathrm{target}}
+
\lambda_{\mathrm{s}} \mathcal{L}_{\mathrm{state}}.
\]

\textbf{Multi-Critic.} To model the heterogeneous reward components separately, we use a critic with five value heads. One head estimates the return from the penalty component, and the other four estimate the returns from body position, linear velocity, angular velocity, and rotation tracking, respectively. The heads share a Transformer backbone but each use a separate MLP to predict their corresponding component value. We sum the five predictions to obtain the value of the total reward for policy optimization.

\textbf{Domain Randomization.} During training, we randomize physical properties, actuation parameters, control latency, and external disturbances to improve the robustness of the learned policy and facilitate sim-to-real transfer. The domain randomization settings used are summarized in Table \ref{tab:domain-randomization}.

\begin{table}[htbp]
\centering
\caption{Domain randomization parameters.}
\label{tab:domain-randomization}
\small
\setlength{\tabcolsep}{4pt}

\begin{tabular}{lll}
\toprule
\textbf{Domain Rand Term} & \textbf{Range} & \textbf{Type} \\
\midrule
Friction
    & $[0.3, 1.6]$
    & Scale \\
Body link mass
    & $[0.5, 1.5]$
    & Scale \\
Torso link mass (kg)
    & $[0, 2]$
    & Add \\
Body link CoM (m)
    & $[-0.03, 0.03]$
    & Add \\
Joint stiffness
    & $[0.75, 1.25]$
    & Scale \\
Action delay
    & $[0, 3]$
    & -- \\
Hands link mass (kg)
    & $[0, 1.2]$
    & Add \\
Torso link CoM (m)
    & $[-0.05, 0.05]$
    & Add \\
Joint armature
    & $[0.75, 1.25]$
    & Scale \\
Joint damping
    & $[0.75, 1.25]$
    & Scale \\
Push robot base (m/s)
    & $[-0.25, 0.25]$
    & Add \\
Push robot hands (N)
    & $[-2.0, 2.0]$
    & Add \\
Push robot feet (N)
    & $[-3.0, 3.0]$
    & Add \\

\bottomrule
\end{tabular}
\end{table}

\textbf{Training Hyperparameters.} We list the training hyperparameters in Table \ref{tab:training-hyperparameters}, including the rollout length, PPO coefficients, optimization settings, learning rate, desired KL divergence, and initial actor standard deviation.

\begin{table}[htbp]
    \centering
    \caption{Training hyperparameters.}
    \label{tab:training-hyperparameters}
    \small
    \setlength{\tabcolsep}{5pt}
    \renewcommand{\arraystretch}{1.03}
    \begin{tabular}{ll}
        \toprule
        \textbf{Training hyperparameter} & \textbf{Value} \\
        \midrule
        Num steps per env                      & 16 \\
        Num learning epochs                        & 1 \\
        Num mini-batches                       & 1 \\
        Gamma                                  & 0.97 \\
        Lambda                                 & 0.9 \\
        Clip parameter                         & 0.2 \\
        Entropy coefficient                    & 0.01 \\
        Value loss coefficient                 & 1.0 \\
        Learning rate                    & $1\times10^{-5}$ \\
        Max gradient norm                      & 1.0 \\
        Desired KL                             & 0.01 \\
        Init actor std                         & 0.5 \\
        
        \bottomrule
    \end{tabular}
    \vspace{4pt}
\end{table}

\subsection{Data Processing Details}

\begin{figure}[H]
    \centering
    \includegraphics[width=\linewidth]{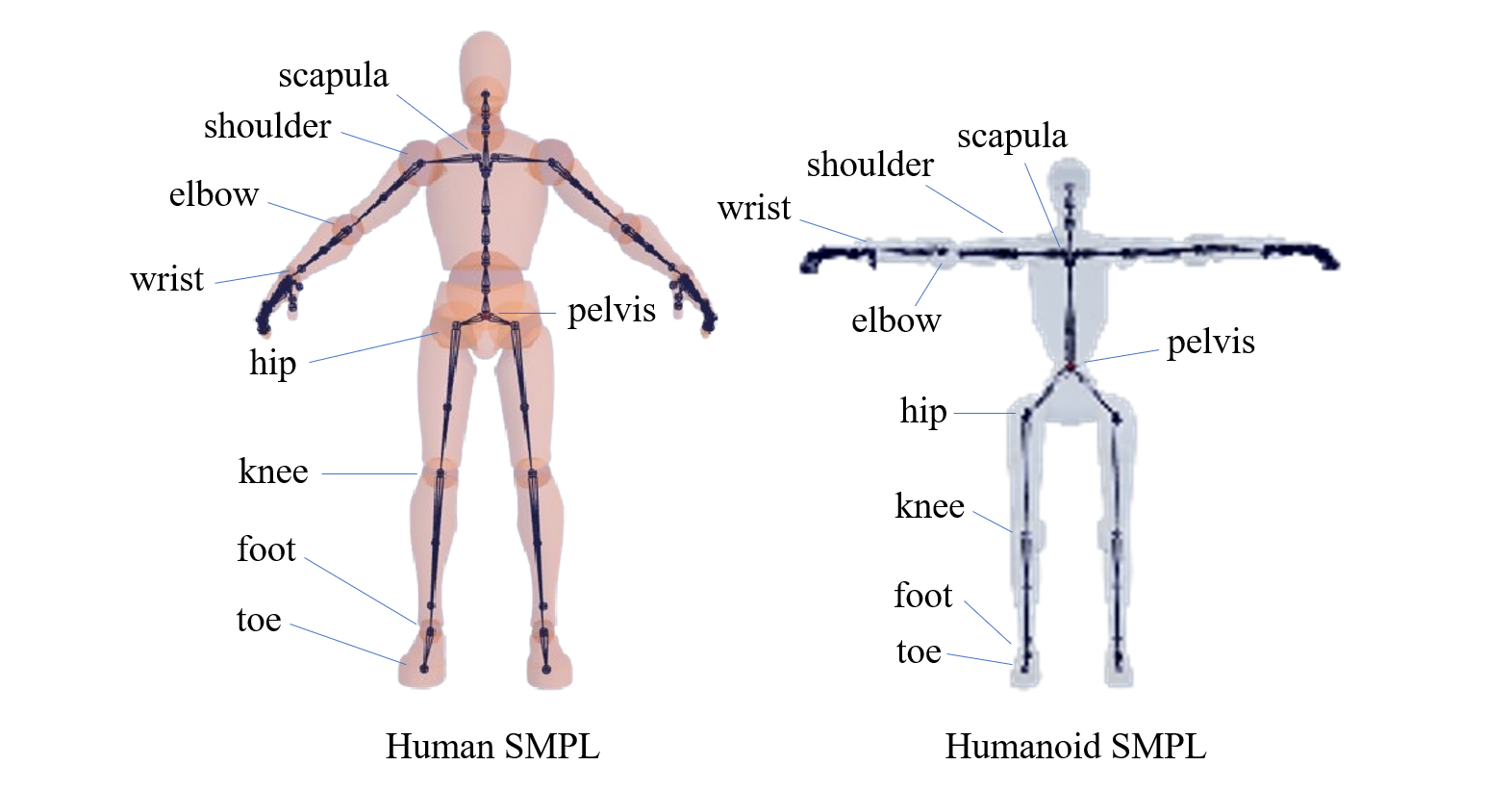}

    \caption{SMPL body models used for shape fitting.}
    \label{fig:two-smpl}
\end{figure}

\textbf{Motion Skeletons.} In the construction of the motion dataset, we first use the human SMPL skeleton to standardize all collected human motions. This step converts heterogeneous motion data from different sources into a unified skeletal representation. After standardization, we perform shape fitting to map the human motions onto a humanoid SMPL skeleton, which is designed to match the morphology of the humanoid robot. The resulting humanoid motions are used as the reference targets for training the generator policy. The SMPL body models are shown in Fig.~\ref{fig:two-smpl}.

\textbf{Augmentation Strategy.} To further improve the diversity and coverage of the training data, we apply three motion augmentation strategies. In addition to the commonly used \textit{mirroring}, we further introduce \textit{stitching} and \textit{recomposition}, which are detailed below.

\subsubsection{Stitch Augmentation} 
\begin{figure}[H]
    \centering
    \includegraphics[width=\linewidth]{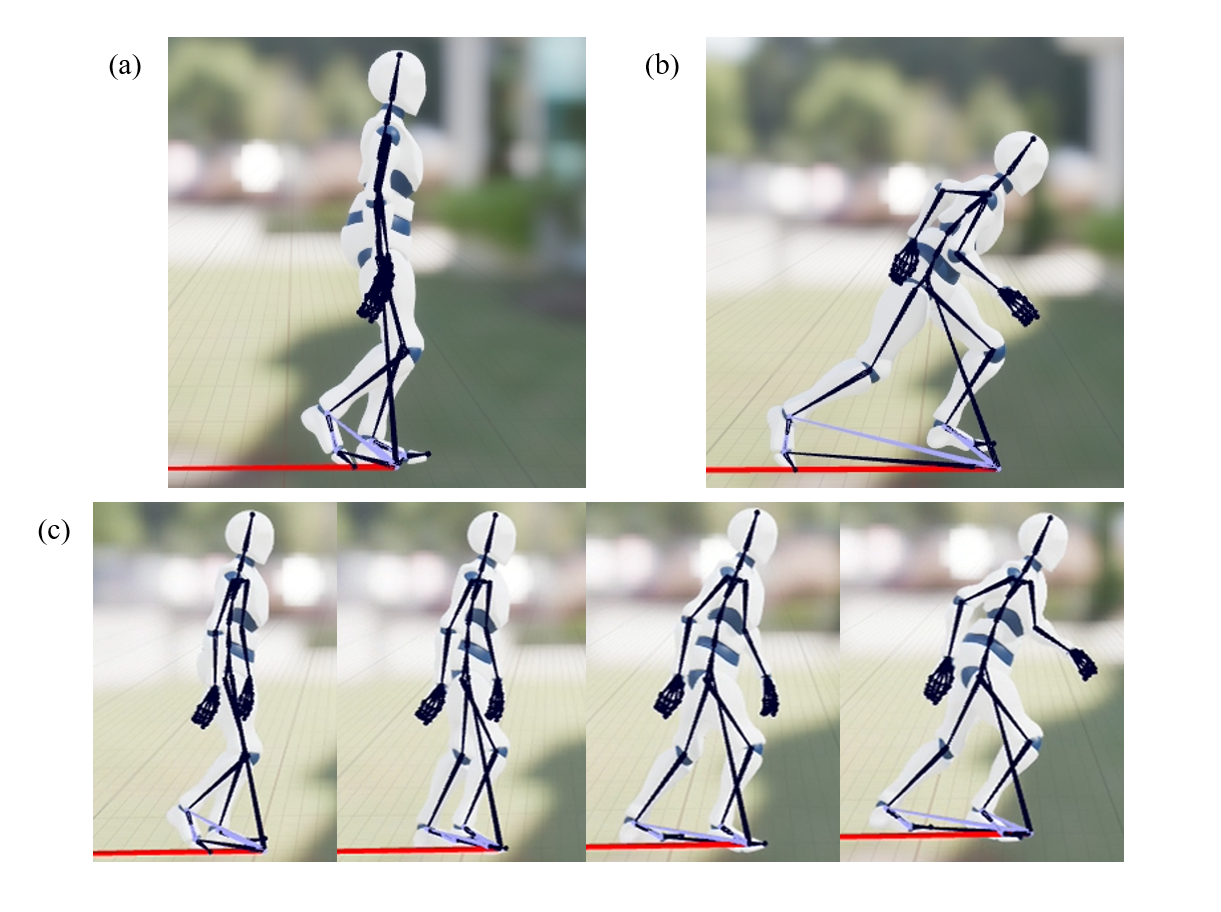}
    \caption{Stitch on motions. (a) Walking; (b) Running;     
    (c) Transition from walking to running.}
    \vspace{-2pt}
    \label{fig:aug:stitch}
\end{figure}

Stitching refers to connecting motion clips from different sequences. Specifically, we identify a pair of frames with similar body configurations from two motion clips and interpolate between them over a short transition window. As illustrated in Fig.~\ref{fig:aug:stitch}, the synthesized transition smoothly connects two originally independent motions, e.g., from walking to running, thereby creating motion sequences with novel temporal compositions.

Beyond increasing motion diversity, stitching also improves the controller's robustness to interpolated motion targets. In practical applications, motion targets may sometimes be specified only at sparse keyframes and then interpolated into continuous trajectories before being fed to the controller. Such interpolated targets can exhibit temporal characteristics different from mocap motions. By explicitly introducing similar interpolation patterns during training, the policy becomes better adapted to this target distribution and can produce smoother and more stable control when driven by keyframe-interpolated trajectories at deployment time.

\subsubsection{Recomposition Augmentation} 
\begin{figure}[H]
    \centering
    \includegraphics[width=\linewidth]{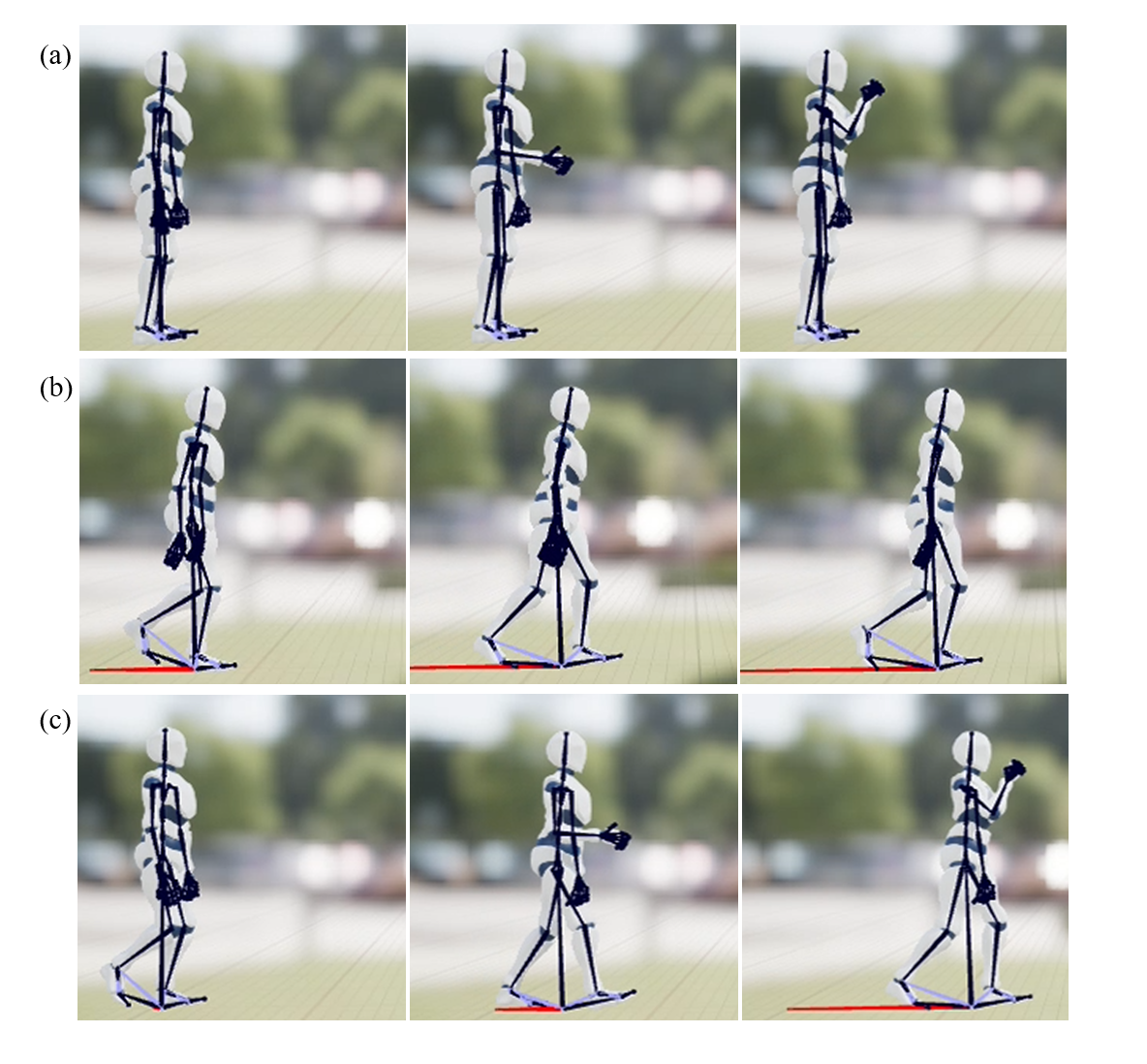}
    \caption{Recomposition on motions. (a) Drinking; (b) Walking; (c) Drinking while walking.}
    \label{fig:aug:recom}
    \vspace{-2pt}
\end{figure}

\textit{Recomposition} refers to combining motions of different body parts from different source sequences. For example, as illustrated in Fig.~\ref{fig:aug:recom}, the upper-body drinking motion from a standing sequence can be combined with the lower-body motion from a walking sequence, yielding a new motion of walking while drinking. Such recomposition is feasible because motions of different body regions can often be treated as relatively independent at the kinematic level. The resulting motion can introduce new whole-body coordination requirements. In this example, lower-body locomotion continuously perturbs the stability of the upper body, while changes in the upper-body configuration in turn affect balance and lower-body locomotion. Successfully executing such a motion requires the controller to coordinate the entire body to resolve these mutual interactions and satisfy the new dynamic constraints introduced by recomposition. Therefore, recomposition not only enlarges the space of motion targets, but also provides challenging training samples that encourage the policy to learn whole-body coordination across motion components that do not originally co-occur in the dataset.

\subsection{Additional Real-World Demonstrations}

Beyond the evaluations in the main paper, we further conduct a series of real-world demonstrations to show the generalization, deployment flexibility, and robustness of GAE under diverse practical conditions.

\begin{figure}[H]
    \centering
    \includegraphics[width=\linewidth]{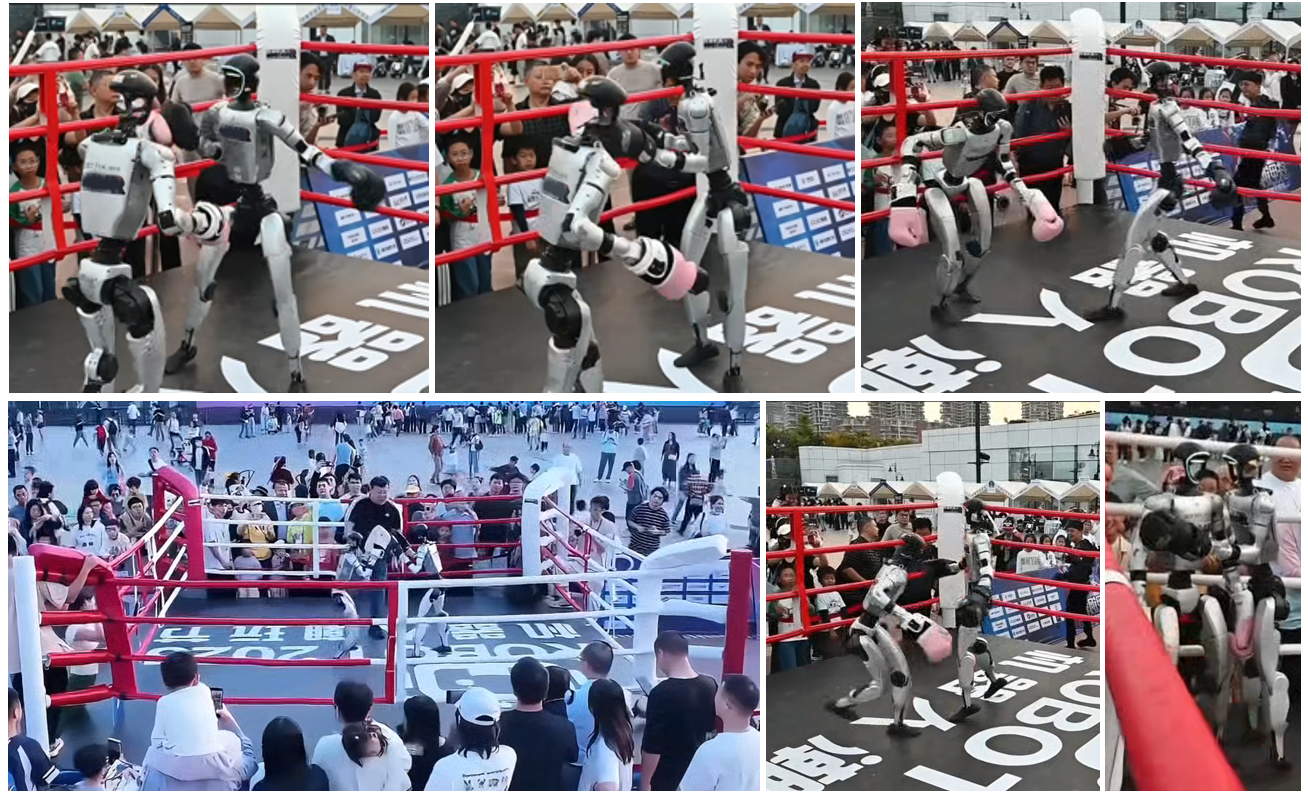}
    \caption{Open-world teleoperation. Two Unitree G1 robots follow separate motion-capture operators during a live fighting performance, reproducing expressive behaviors.}
    \label{fig:fight}
\end{figure}

\textbf{Open-ended Fight Performance.} GAE was deployed in publicly held live performance events. As shown in Fig.\ref{fig:fight}, two Unitree G1 robots controlled by GAE were teleoperated by separate motion-capture operators during a live fighting performance. Unlike controlled laboratory trials, this scenario involves unscripted operator behaviors, rapidly changing whole-body motions, and long-duration execution in a realistic environment. During the performance, the robots reproduced a wide range of movements, including fast arm swings, turns, steps, and kicks, while remaining stable throughout the demonstration. This deployment illustrates GAE's potential for real-time humanoid teleoperation beyond scripted benchmark motions.

\begin{figure}[H]
    \centering
    \includegraphics[width=\linewidth]{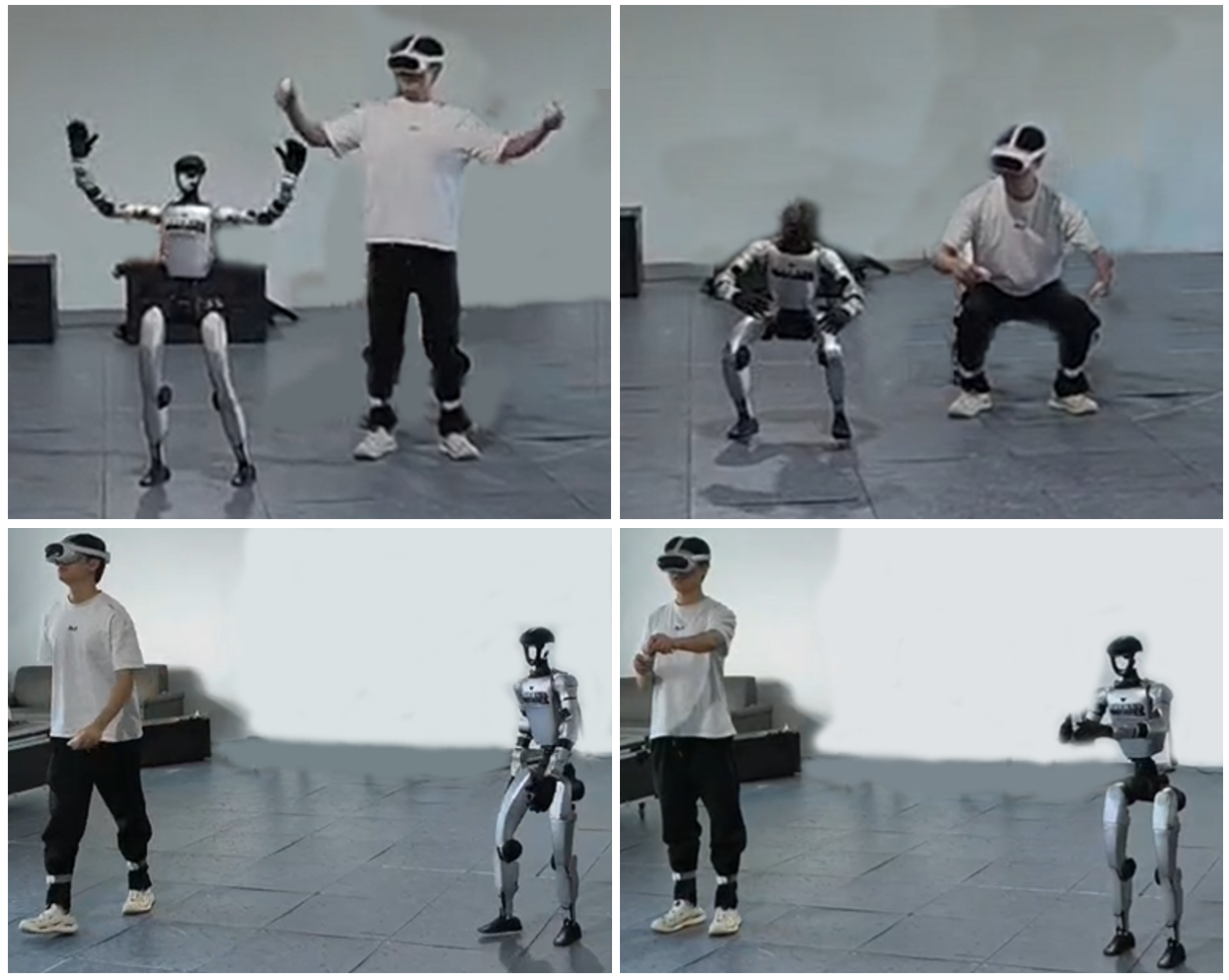}
    \caption{VR-based teleoperatopn. The operator's head, hands, and feet are tracked by a VR system and reconstructed into whole-body motion. GAE follows the reconstructed motion to produce smooth robot movements.}
    \label{fig:vr}
\end{figure}

\textbf{VR-Based Teleoperation.} In VR-based teleoperation setup, the operator's movements are captured through sparse end-effector tracking, including head, hands, and feet. The built-in motion reconstruction system of the VR platform can estimate a whole-body SMPL motion sequence from these sparse observations, which can be directly converted into the motion target required by GAE. As shown in Fig.~\ref{fig:vr}, GAE can effectively track the reconstructed whole-body motion target and produce smooth behaviors, demonstrating its compatibility with VR-based teleoperation interfaces.

\begin{figure}[H]
    \centering
    \includegraphics[width=\linewidth]{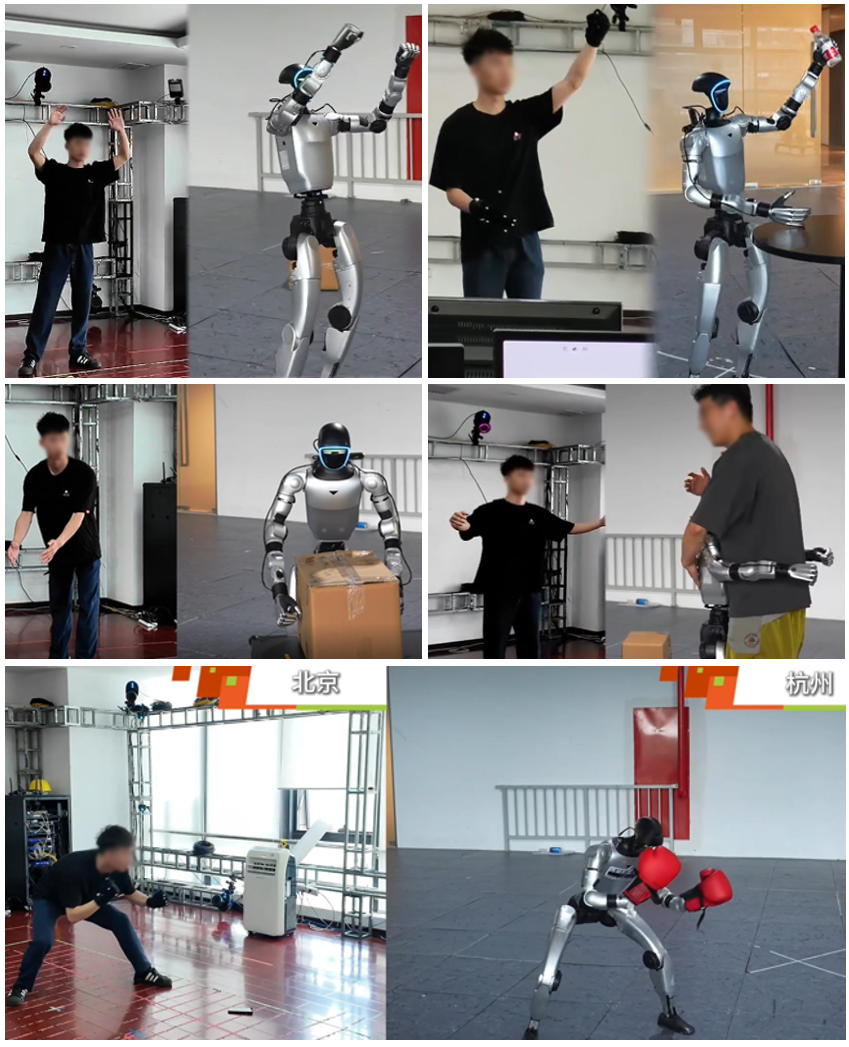}
    \caption{Long-distance teleoperation. An operator in Beijing remotely controls the humanoid robot in Hangzhou over a distance of approximately 1,300 km.}
    \label{fig:remote}
    \vspace{-2pt}
\end{figure}

\textbf{Long-Distance Teleoperation.} We evaluated GAE in a long-distance teleoperation experiment spanning approximately 1,300 km, with the operator in Beijing and the humanoid robot in Hangzhou. Compared with local teleoperation, this setting introduces more challenging communication conditions and places higher demands on the synchronization and robustness of the control system. To mitigate the effects of network latency, we enabled GAE's latency-conditioned anticipation during remote operation. As shown in Fig.~\ref{fig:remote}, GAE enables the robot to closely follow the operator through arm waving, stepping, and boxing motions. It also uses its dexterous hands to grasp a bottle and handle a cardboard box, demonstrating object interaction alongside whole-body teleoperation. Throughout the demonstration, the robot maintains stable and smooth control despite the geographical separation.

\begin{figure}[H]
    \centering
    \includegraphics[width=\linewidth]{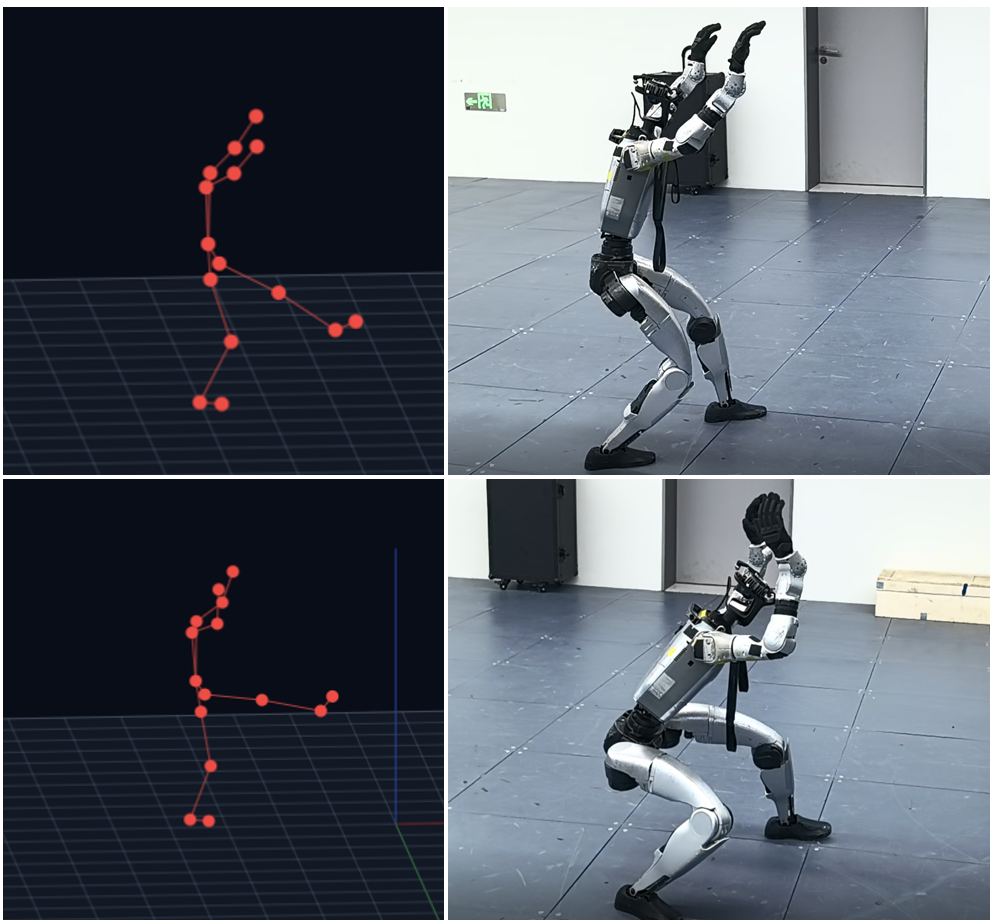}
    \caption{Robustness to command interruption.  When the motion target is unexpectedly frozen at an unstable frame(left), GAE autonomously settles the robot into a nearby balanced posture(right) without falling or losing control.}
    \label{fig:interrupt}
\end{figure}

\textbf{Robustness to Command Interruption.} We evaluate GAE under unexpected interruptions of the motion target, which may occur in practice due to network failures or temporary disconnection of the motion-capture device. Once the motion stream is interrupted, the robot continuously receives the last available motion target, which may correspond to an unstable configuration, such as a kicking motion with one leg suspended in the air. As shown in Fig.~\ref{fig:interrupt}, rather than rigidly maintaining such an infeasible target and becoming unstable, GAE autonomously settles into a nearby balanced posture while remaining close to the last target motion. This behavior prevents the robot from falling or losing control under command interruption, improving the safety and robustness of teleoperation in practical deployment.

\begin{figure}[H]
    \centering
    \includegraphics[width=\linewidth]{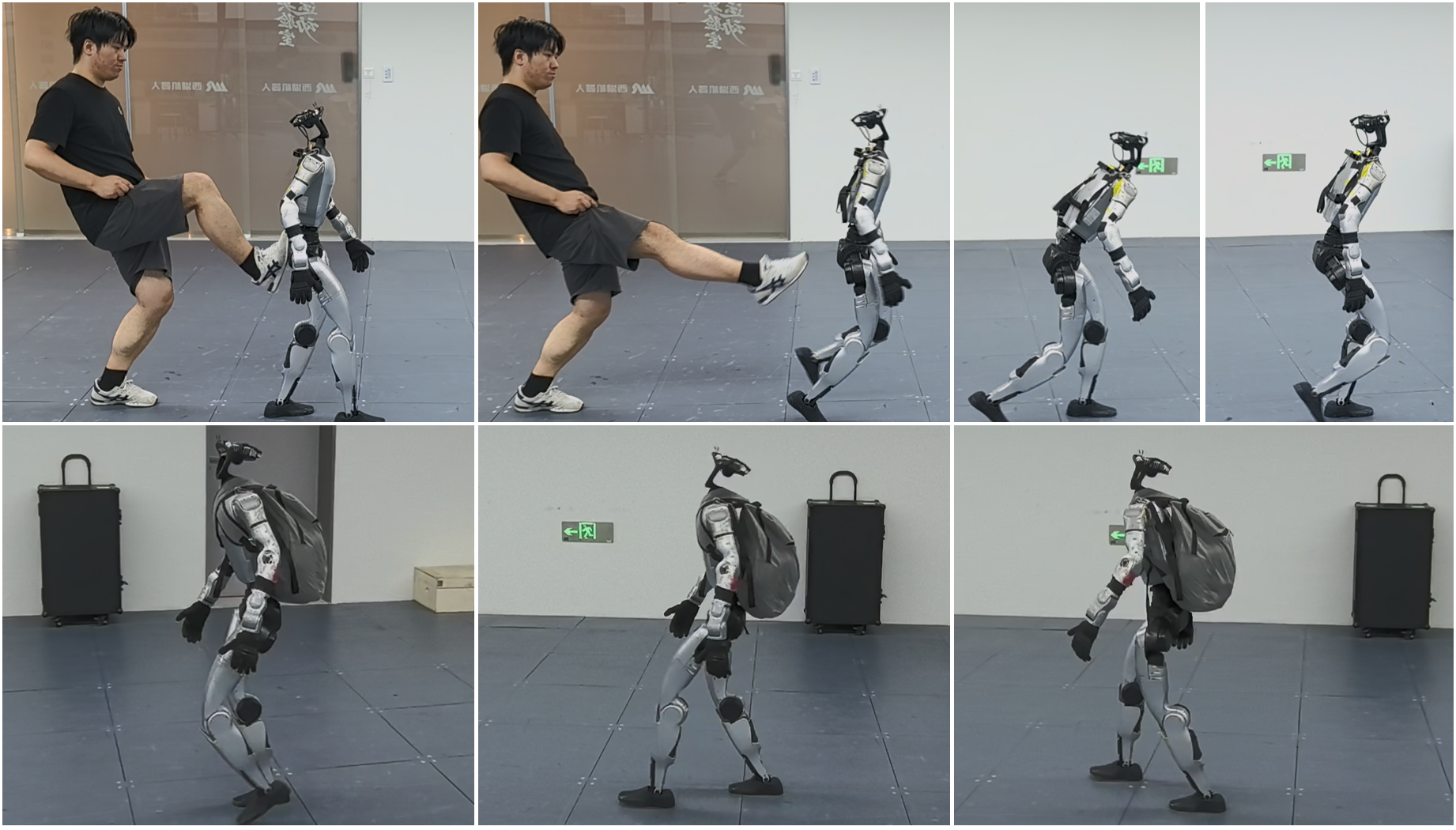}
    \caption{Robustness to physical disturbances. GAE maintains stable whole-body control under a strong external impact to the pelvis and an additional payload of approximately 5kg.}
    \label{fig:push}
\end{figure}

\textbf{Robustness to Physical Disturbances.} We further evaluate the robustness of GAE under strong external disturbances and additional payloads. As shown in Fig.~\ref{fig:push}, we first apply a sudden and strong impact to the pelvis of the robot to mimic an unexpected collision. Although the robot staggers forward for several steps after the impact, it is able to recover its balance and continue stable operation. We further attach an approximately 5kg payload to the robot to evaluate its behavior under additional loading. Despite the change in body dynamics, the robot remains stable and continues to execute smooth whole-body motions, demonstrating the robustness of GAE to both transient external disturbances and persistent payload variations.

\begin{figure}[H]
    \centering
    \includegraphics[width=\linewidth]{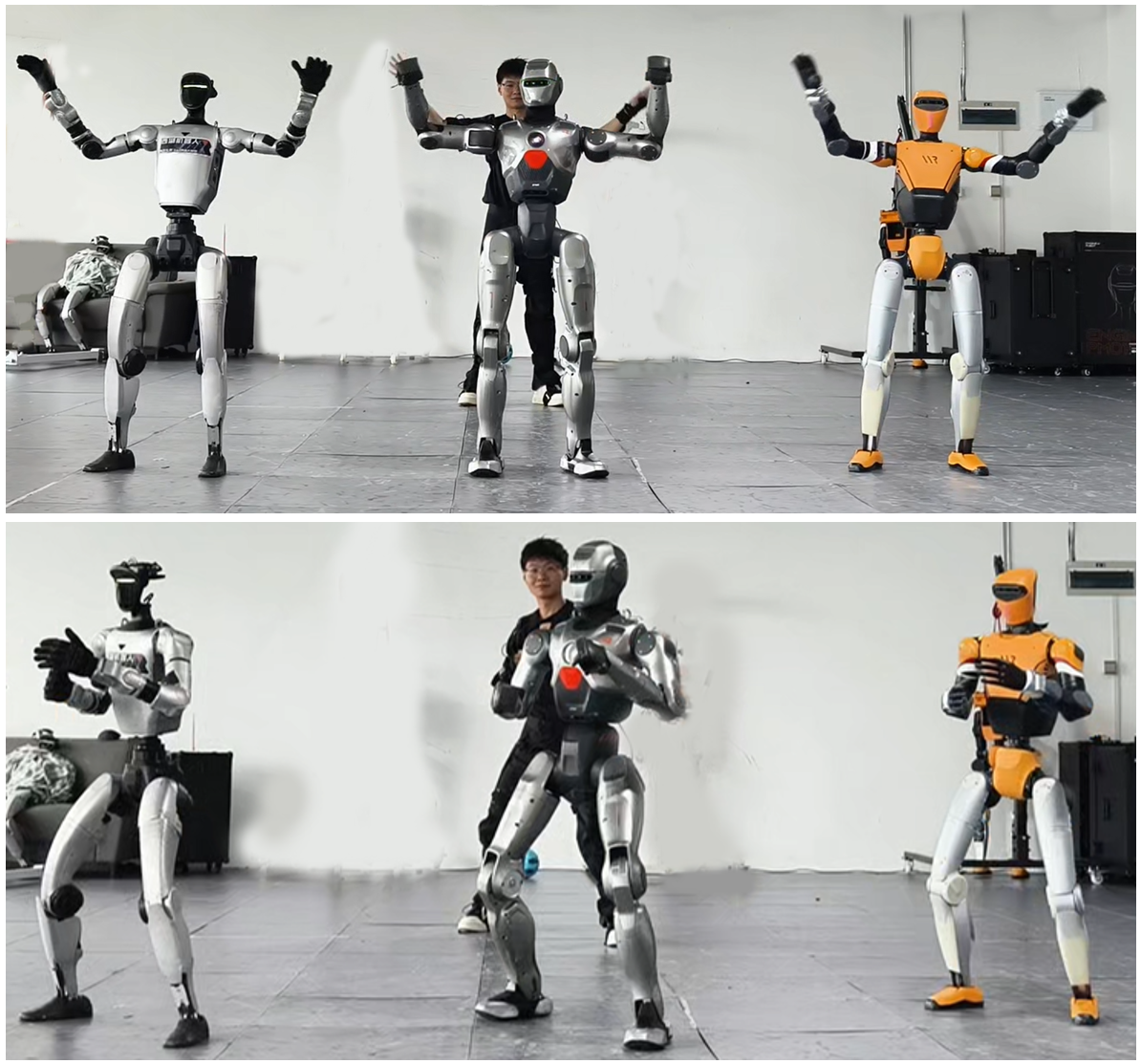}
    \caption{Cross-embodiment teleoperation. Unitree G1, Westlake O1, and ENGINEAI PM01 are simultaneously controlled by the same human operator, demonstrating the transferability of the GAE framework across different humanoid embodiments.}
    \label{fig:cross}
    \vspace{-2pt}
\end{figure}

\textbf{Cross-Embodiment Transfer.} In the main paper, we have demonstrated the transferability of the GAE framework from Unitree G1 to Westlake O1. Here, we further extend the evaluation to another humanoid platform, the ENGINEAI PM01. As shown in Fig.~\ref{fig:cross}, Unitree G1, Westlake O1, and ENGINEAI PM01 are controlled simultaneously by the same human operator and execute coordinated whole-body motions. Despite their differences in morphology and kinematic structure, all three robots are able to follow the operator effectively, further demonstrating that the GAE framework can be transferred across diverse humanoid embodiments.

\end{document}